\documentclass[10pt,twocolumn,letterpaper]{article}

\usepackage{cvpr}              

\usepackage{graphicx}
\usepackage{amsmath}
\usepackage{amssymb}
\usepackage{booktabs}

\usepackage[pagebackref,breaklinks,colorlinks]{hyperref}

\usepackage[capitalize]{cleveref}
\crefname{section}{Sec.}{Secs.}
\Crefname{section}{Section}{Sections}
\Crefname{table}{Table}{Tables}
\crefname{table}{Tab.}{Tabs.}

\def\cvprPaperID{5389} 
\def\confName{CVPR}
\def\confYear{2023}

\usepackage{amsmath,amsfonts,bm}

\def\eqref#1{equation~\ref{#1}}

\def\1{\bm{1}}

\DeclareMathAlphabet{\mathsfit}{\encodingdefault}{\sfdefault}{m}{sl}
\SetMathAlphabet{\mathsfit}{bold}{\encodingdefault}{\sfdefault}{bx}{n}

\DeclareMathOperator*{\argmin}{arg\,min}

\usepackage[utf8]{inputenc} 
\usepackage[T1]{fontenc}    
\usepackage{hyperref}       
\hypersetup{breaklinks=true,colorlinks,citecolor=blue}
\usepackage{url}            
\usepackage{booktabs}       
\usepackage{amsfonts}       
\usepackage{nicefrac}       
\usepackage{microtype}      
\usepackage[table,xcdraw]{xcolor}
\usepackage{subcaption}
\usepackage{bm}
\usepackage{wrapfig}
\usepackage{amsmath}
\usepackage{multirow}
\usepackage{graphicx,paralist,booktabs,tabu}
 
\newcommand{\topic}[1]{\vspace{1mm}\noindent\textbf{#1}}

\newcommand{\norm}[1]{\left\lVert#1\right\rVert}
\usepackage{makecell}

\usepackage{amsthm}

\newcommand{\quotes}[1]{``#1''}
\newcommand{\Poincare}{Poincar\'e\xspace}
\usepackage{enumitem}
\newcommand{\bfsection}[1]{\vspace*{0.1cm}\noindent\textbf{#1}}

\title{Hyperbolic Contrastive Learning}

\begin{document}

\author{Yun Yue\textsuperscript{\rm 1},
Fangzhou Lin\textsuperscript{\rm 1,2},
Kazunori D Yamada\textsuperscript{\rm 2},\\
Ziming Zhang\textsuperscript{\rm 1}
\\
\textsuperscript{\rm 1}Worcester Polytechnic Institute, Worcester, USA,
\textsuperscript{\rm 2}Tohoku University, Sendai, Japan\\
\texttt{\{yyue,zzhang15\}@wpi.edu} 
}

\maketitle

\begin{abstract}
Learning good image representations that are beneficial to downstream tasks is a challenging task in computer vision. As such, a wide variety of self-supervised learning approaches have been proposed. Among them, contrastive learning has shown competitive performance on several benchmark datasets. The embeddings of contrastive learning are arranged on a hypersphere that results in using the inner (dot) product as a distance measurement in Euclidean space. However, the underlying structure of many scientific fields like social networks, brain imaging, and computer graphics data exhibit highly non-Euclidean latent geometry. We propose a novel contrastive learning framework to learn semantic relationships in the hyperbolic space. Hyperbolic space is a continuous version of trees that naturally owns the ability to model hierarchical structures and is thus beneficial for efficient contrastive representation learning. We also extend the proposed \emph{Hyperbolic Contrastive Learning (HCL)} to the supervised domain and studied the adversarial robustness of \emph{HCL}. The comprehensive experiments show that our proposed method achieves better results on self-supervised pretraining, supervised classification, and higher robust accuracy than baseline methods.
\end{abstract}

\section{Introduction}


In computer vision, downstream tasks could be fine-tuned efficiently and effectively with good image representations. However, learning a good image representation remains a challenging task \cite{fukushima1982neocognitron, wiskott2002slow, hinton2006fast, oquab2014learning, simonyan2014very, girshick2014rich, long2015fully}. The \quotes{pretext} self-supervised learning \cite{trinh2019selfie, noroozi2016unsupervised, carlucci2019domain, gidaris2018unsupervised} relies on heuristic handcrafted task design to learn representations.
Recently, contrastive learning \cite{chopra2005learning, hadsell2006dimensionality} has become the dominant method in self-supervised learning and has shown competitive performance over its supervised counterpart on several downstream tasks such as classification, object detection, and segmentation \cite{oord2018representation, jaiswal2020survey, deng2009large, misra2020self, he2020momentum, everingham2010pascal, guler2018densepose, he2017mask, lin2014microsoft, faster2015towards}. 

Typically, given an anchor point $\mathbf x$, contrastive learning takes augmented views of the same data as positive pairs $(\mathbf x, \mathbf x^+)$, and other data in the same batch as negative pairs $(\mathbf x, \mathbf x^-)$. Since the similarity in the embedding space reflects the similarity of semantics. The contrastive representation learning attempts to pull the embeddings of positive pairs closer and push the embeddings of negative pairs away in the latent space by optimizing the objective such as the InfoNCE loss \cite{oord2018representation, chen2020simple}.

Despite the promising results shown by current contrastive learning literature, it suffers from a fundamental limitation that has been encountered by many embedding methods: the ability to model complex patterns is inherently bounded by the dimensionality of the embedding space \cite{nickel2017poincare}. The embeddings of contrastive learning are arranged on a hypersphere that results in using inner (dot) product as a distance measurement. However, the underlying structure of many scientific fields like social networks, brain imaging, and computer graphics data are hierarchical \cite{bronstein2017geometric}. In this paper, we attempt to build an efficient learning framework by introducing hyperbolic space.

Different from the Euclidean space $\mathbb{R}^n$ that has polynomial volume growth \wrt{}the radius, the hyperbolic space $\mathbb{H}^n$ has exponential growth that is suitable for tree-like structure data. The representation power of hyperbolic space has been demonstrated in NLP \cite{nickel2017poincare, nickel2018learning} as well as image segmentation \cite{weng2021unsupervised, atigh2022hyperbolic}, few-shot \cite{khrulkov2020hyperbolic} and zero-shot learning \cite{liu2020hyperbolic}, and metric learning equipped with vision transformers \cite{ermolov2022hyperbolic}. To better reveal the underlying hierarchical structure of data, we explore the potential of hyperbolic space, where the curvature is a constant negative, in contrastive learning. Instead of computing the feature similarity in Euclidean space, we project data to hyperbolic space for distance measurement. Similar to the general tree structure, the hyperbolic space is a continuous version of trees that naturally owns the ability to model hierarchical structures and is thus beneficial for efficient contrastive learning.  

\begin{itemize}[nosep, leftmargin=*]
\item In this paper, we propose \emph {Hyperbolic Contrastive Learning (HCL)}, a new contrastive learning framework for self-supervised image pretraining that leverages the representation power of hyperbolic space. 
 
\item We further propose \emph {Supervised Hyperbolic Contrastive Learning (SHCL)}, a variant of general supervised contrastive loss that yields even better performance on supervised image classification tasks.
\end{itemize}
Though the feature consistency \wrt{}data augmentations introduced by the contrastive loss is effective for standard generalization of CNNs and unsupervised learning \cite{zhang2019making, kayhan2020translation, hu2017learning, ye2019unsupervised, xie2020unsupervised, jiang2020robust}, deep learning models 
also exhibit adversarial fragility \cite{chen2020adversarial}. Considering pretrained models from self-supervision is usually used in downstream tasks for faster fine-tuning or better accuracy, it is natural to explore whether pretrained self-supervised models play a similar role for adversarial training as they have for standard training \cite{chen2020adversarial}. The connection between self-supervised contrastive learning and adversarial training has not been built until recently \cite{jiang2020robust, kim2020adversarial}. 
\begin{itemize}[nosep, leftmargin=*]
\item To develop label-efficient and robust models, we further investigate \emph {Robust Hyperbolic Contrastive Learning (RHCL)} and verify that hyperbolic space is more suitable for contrastive learning and more robust to adversarial attacks.
\end{itemize}
Our approach is based on \Poincare model, a particular model of hyperbolic space that is well-suited for gradient-based optimization \cite{nickel2017poincare}. To the best of our knowledge, we are the first to investigate contrastive learning, its supervised counterpart, as well as its adversarial robustness in the hyperbolic space. Empirically, We show that the proposed \emph{HCL} and \emph{SHCL} are more appropriate in capturing the underlying relationships of image data and thus result in better classification performance on several benchmark datasets. We also demonstrate that \emph{RHCL} is more robust to adversarial perturbations compared with general contrastive learning and other baseline methods.

\vspace{-0.2cm}
\section{Related Work}

\topic{Self-supervised Contrastive Learning.}
The idea of pulling the representations of similar data (positive pairs) in the embedding space while pushing away representations of dissimilar data (negative pairs) has long been the classic idea in metric learning \cite{musgrave2020metric}. One of the fundamental losses in metric learning is contrastive loss \cite{hadsell2006dimensionality}. A wide variety of its variants has since been proposed \cite{weinberger2009distance, oh2016deep, goldberger2004neighbourhood, sohn2016improved, boudiaf2020unifying}. Recently, learning representations from unlabeled data in contrastive way \cite{chopra2005learning, hadsell2006dimensionality} has been one of the most competitive research field \cite{oord2018representation, hjelm2018learning, wu2018unsupervised, tian2020contrastive, sohn2016improved, chen2020simple, jaiswal2020survey, li2020prototypical, he2020momentum, chen2020improved, chen2020big, bachman2019learning, misra2020self, caron2020unsupervised}. Popular model structures like SimCLR \cite{chen2020simple} and Moco \cite{he2020momentum} apply the commonly used loss function InfoNCE \cite{oord2018representation} to learn a latent representation that is beneficial to downstream tasks. Several theoretical studies show that self-supervised contrastive loss optimizes data representations by aligning the same image's two views (positive pairs) while pushing different images (negative pairs) away on the hypersphere \cite{wang2020understanding, chen2021intriguing, wang2021understanding, arora2019theoretical}. Though these pair-based methods in self-supervised contrastive learning do not require labels, they rely heavily on calculating Euclidean distance between data embeddings. In addition, hierarchical semantic structures naturally exist in image datasets. While other work like the Hierarchical Contrastive Selective Coding (HCSC)  \cite{guo2022hcsc} learns a set of hierarchical prototypes to represent the hierarchical semantic structures underlying the data in the latent space explicitly, our work proposes to learn the hierarchy structure in a different space with hyperbolic embedding without defining extra prototypes.

\topic{Adversarial Robustness of Contrastive Learning.}
Deep Neural Networks (DNNs) for computer vision are vulnerable to small image perturbations \cite{szegedy2013intriguing, goodfellow2014explaining, carlini2017towards}. For example, small perturbations to the visual input can result in large feature variations and crucial challenges in safety-critical applications \cite{szegedy2013intriguing, carlini2017towards, goodfellow2014explaining, moosavi2017universal, papernot2016limitations, eykholt2018robust}. Adversarial defense algorithms have been proposed in response to the adversarial threat \cite{zhang2019theoretically, qin2019adversarial, madry2017towards}. One of the most popular approaches is adversarial training (AT) \cite{madry2017towards}, which trains the neural network with the worst-case adversarial examples.
Although existing contrastive learning literature has shown boosted performance on the standard generalization, its connection with adversarial robustness has not been studied until recently \cite{kim2020adversarial, jiang2020robust}. For detailed review please refer \cite{qi2022adversarial}. Essentially, SimCLR encourages feature consistency to specified data augmentations. Coincidentally, enforcing consistency during training \wrt{} perturbations has been shown to smooth feature space near samples and thus immediately help adversarial robustness \cite{alayrac2019labels, zhai2019adversarially, carmon2019unlabeled}. Several closely relevant works have investigated improving the adversarial robustness via self-supervised contrastive pretraining \cite{chen2020adversarial, kim2020adversarial, yu2022adversarial, gupta2022higher}.
We draw upon the insights in these works and seek to investigate the adversarial robustness of contrastive learning in a new embedding space.

\topic{Supervised Contrastive Learning.}
The cross-entropy loss has been the most widely used loss function for supervised learning of deep classification models for years. With the development of contrastive learning, the approach has been extended to fully-supervised setting \cite{khosla2020supervised}. Since the label information is known in supervised setting, instances from the same class naturally form a positive data pool, and data from different classes are negatives. Technically, each anchor has many positives as opposed to self-supervised contrastive learning which uses only a single augmentation as a positive. The supervised contrastive (SupCon) loss has been shown consistently outperforms cross-entropy on several image classification tasks \cite{khosla2020supervised}. Our work investigates the hyperbolic space application on both self-supervised domain and its supervised extension which will provide preliminary results for some future studies in this direction.  

\topic{Hyperbolic Embeddings.}
The Euclidean space has been widely used by the machine learning community for representation learning as this space is a natural generalization of our intuition-friendly, visual three-dimensional space, and the easy measurement of distance with inner-product in this space \cite{ganea2018hyperbolic, peng2021hyperbolic}. However, the Euclidean embedding is not the suitable choice for some complex tree-like data fields such as Biology, Network Science, Computer Graphics, or Computer Vision that exhibit highly non-Euclidean latent geometry \cite{ganea2018hyperbolic, bronstein2017geometric}. This encourages the research community to develop deep neural networks in non-Euclidean space such as the hyperbolic space, which is a Riemannian manifold of constant negative curvature. Recently, the gap between the hyperbolic embeddings and the Euclidean embeddings has been narrowed by deriving the essential components of deep neural networks in hyperbolic geometry (\eg{multinomial logistic regression, fully-connected layers, and recurrent neural networks \etc{}}). \cite{ganea2018hyperbolic, shimizu2020hyperbolic}. In the computer vision domain, the hyperbolic space has been found well-suited for image segmentation \cite{weng2021unsupervised, atigh2022hyperbolic}, zero-shot recognition \cite{liu2020hyperbolic, fang2021kernel}, few-shot image classification \cite{khrulkov2020hyperbolic, fang2021kernel, gao2021curvature} as well as point cloud classification \cite{montanaro2022rethinking}. The work of \cite{guo2022clipped} revealed the vanishing gradients issue of Hyperbolic Neural Networks (HNNs) when applied to classification benchmarks that may not exhibit hierarchies and showed clipped HNNs are more robust to adversarial attacks. 
Concurrently, \cite{ermolov2022hyperbolic} mapped the output of image representations encoded by a vision transformer and a fully-connected layer to a hyperbolic space to group the representations of similar objects in the embedding space. Their goal is to investigate the pairwise cross-entropy loss with the hyperbolic distance function in metric learning while we focus on investigating the self-supervised and supervised contrastive learning in hyperbolic space.

\section{Method}
We briefly introduce the hyperbolic geometry and \Poincare embedding in Section \ref{sec:method-hyper}. We then discuss the proposed frameworks HCL and SHCL in Section \ref{sec:method-info} and \ref{sec:method-sup}. We describe the adversarial robustness of HCL in Section \ref{sec:method-adv}.

\subsection{Hyperbolic Geometry \& Embedding}
Unlike the Euclidean geometry where the circle length (\(2\pi r\)) and disc area (\(2\pi r^2\)) grow only linearly
and quadratically with regard to \(r\), hyperbolic disc area and circle length grow exponentially with their
radius \cite{nickel2017poincare}. In hyperbolic geometry, only two dimensions are needed to represent a regular tree with branching
factor \(b\) with \((b + 1)b^{\ell-1}\) nodes at level \(\ell\) and \(((b + 1)b^\ell - 2) /
(b - 1)\) nodes on a level less or equal than \(\ell\). Thus the space is a natural choice for complex data with hierarchical structure.

Following the assumption in general contrastive learning, we are interested in finding an embedding that the distance in the latent space can reflect semantic similarity. In addition, we are interested in embedding the latent hierarchy efficiently. In the meantime,  we do not assume we have access to the hierarchy information of the data. Among several isometric models \cite{cannon1997hyperbolic} of hyperbolic space, similar to \cite{ermolov2022hyperbolic} and \cite{nickel2017poincare}, we stick to the \Poincare ball model \cite{sarkar2012low} that is well-suited for gradient-based optimization (\ie the distance function is differentiable). The embedding is fully unsupervised in our proposed HCL. In particular, the model $(\mathbb{M}^n_c, g^{\mathbb{M}})$ is defined by the manifold $\mathbb{M}^n = \{ x \in \mathbb{R}^n \colon c\|x\|^2 < 1, c \geq 0\} $ equipped with the Riemannian metric $g^{\mathbb{M}} = \lambda_c^2 g^E$, where $c$ is the curvature parameter, $ \lambda_c = \frac{2}{1-c\|x\|^2}$ is conformal factor that scales the local distances and $g^E = \mathbf{I}_n$ denotes the Euclidean metric tensor.  
\begin{figure}
\begin{center}
\includegraphics[width=0.7\linewidth]{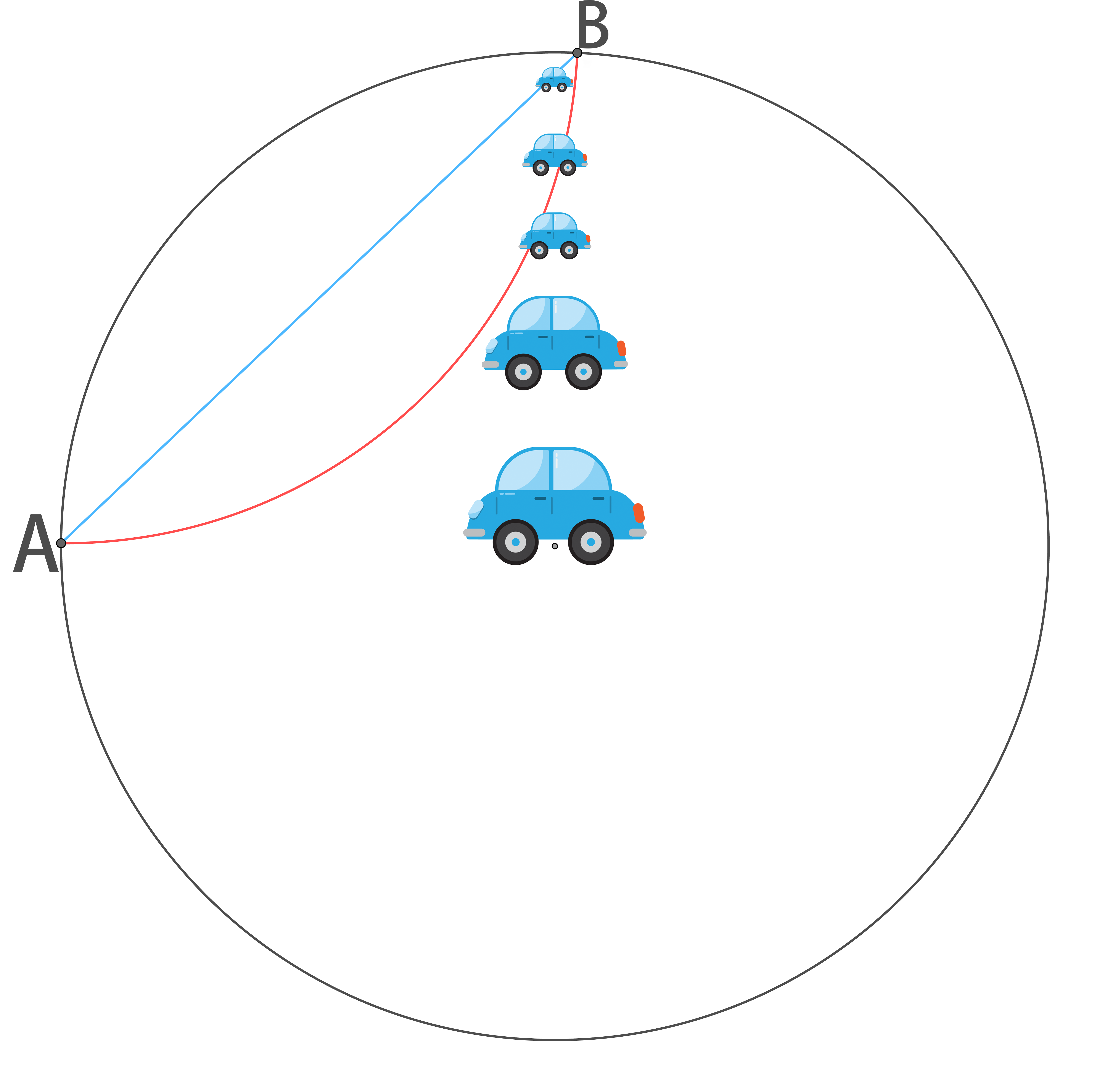}
\vspace{-2mm}
\caption{The geodesics (red curve) of two points ($A, B$) of the \Poincare disk. The blue line is the straight line between $A,B$ that is no longer the shortest distance. The relative size of objects in this disk is getting smaller while the distance of points increases exponentially (relative to their Euclidean distance) when they are getting closer to the boundary.}
\label{fig:disk}
\end{center}
\vspace{-3mm}
\end{figure}

\begin{figure*}[tb]
\vspace{-0.2cm}
\includegraphics[width=1\linewidth]{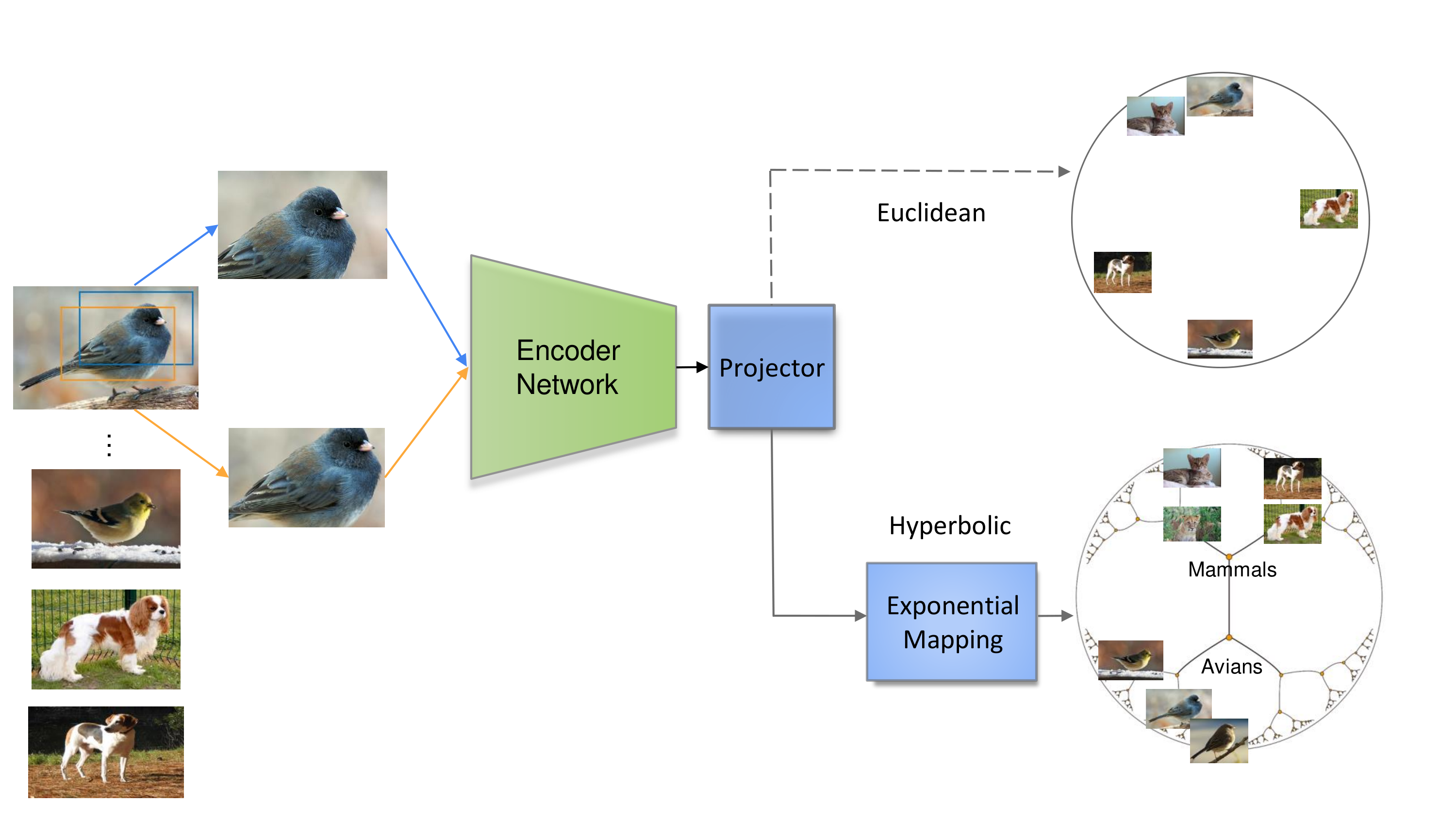}
\caption{Overview of the proposed \textit{HCL}. Different image augmentations are applied to the same image first. After going through the encoder network, the vectors are then projected into a latent space with a fully connected (FC) layer. Different from the general contrastive learning (dashed line) that maps the data to the unit sphere in Euclidean space, our method (solid arrow) maps the data to hyperbolic space. We illustrate the tree embedding in a two-dimensional unit ball in the hyperbolic space. Due to the tree structure representation power of hyperbolic space, our method tends to capture the hierarchy of data while traditional contrastive learning might push the same class of data away since they are treated as negatives. Note class labels are not available during self-supervised pretraining.}
\label{fig:flow}
\vspace{-0.2cm}
\end{figure*}

The framework of \emph{gyrovector spaces} provides an elegant non-associative algebraic formalism for hyperbolic geometry just as vector spaces provide the algebraic setting for Euclidean geometry \cite{cannon1997hyperbolic, ungar2008analytic, ungar2008gyrovector, ganea2018hyperbolic}. For two vectors $\mathbf{x}, \mathbf{y} \in \mathbb{M}^n_c$, their addition is defined as
\begin{equation}\label{eq:add}
    \mathbf{x} \oplus_c \mathbf{y} = \frac{(1+2c\langle \mathbf{x}, \mathbf{y} \rangle + c\|\mathbf{y}\|^2) \mathbf{x}+ (1-c\|\mathbf{x}\|^2)\mathbf{y}}{1+2c\langle \mathbf{x}, \mathbf{y} \rangle + c^2 \|\mathbf{x}\|^2 \|\mathbf{y}\|^2}.
\end{equation}
The hyperbolic distance between $\mathbf{x}, \mathbf{y} \in \mathbb{M}^n_c$ is defined as:
\begin{equation}\label{eq:hdist}
    D_{hyp}(\mathbf{x},\mathbf{y}) = \frac{2}{\sqrt{c}} \mathrm{arctanh}(\sqrt{c}\|-\mathbf{x} \oplus_c \mathbf{y}\|).
\end{equation}
In particular, when $c = 0$, the Eq. \ref{eq:add} is the Euclidean addition of two vectors in $\mathbb{R}^n$ and Eq. \ref{eq:hdist} recovers Euclidean geometry: $\lim_{c \to 0} D_{hyp}(\mathbf{x},\mathbf{y})=2\|\mathbf{x}-\mathbf{y}\|.$ For an open $n$-dimensional unit ball, the geodesics of the \Poincare disk are then circles that are orthogonal to the boundary of the ball. See Fig.
\ref{fig:disk} for an illustration.  

Before performing operations in the hyperbolic space, a bijective map from $\mathbb{R}^n$ to $\mathbb{M}_c^n$ that maps Euclidean vectors to the hyperbolic space is necessary. Such a map is termed \emph{exponential} map when mapping from Euclidean space to the \Poincare model of hyperbolic geometry and the inverse to it is called \emph{logarithmic} map \cite{khrulkov2020hyperbolic}. 

The \emph{exponential} map is defined as:
\begin{equation}\label{eq:exp}
    \exp_\mathbf{x}^c(\mathbf{v}) = \mathbf{x} \oplus_ c \bigg(\tanh \bigg(\sqrt{c} \frac{\lambda_\mathbf{x}^c \|\mathbf{v}\|}{2} \bigg) \frac{\mathbf{v}}{\sqrt{c}\|\mathbf{v}\|}\bigg).
\end{equation}
In practice, we follow the setting of \cite{khrulkov2020hyperbolic} and \cite{ermolov2022hyperbolic} with the base point $\mathbf{x}=\mathbf{0}$ so that the formulas are less cumbersome and empirically have little impact on the obtained results.
\label{sec:method-hyper}

\subsection{Hyperbolic Contrastive Learning}
\label{sec:method-info}
Fig. \ref{fig:flow} shows our proposed framework Hyperbolic Contrastive Learning. We follow the name convention of \cite{khosla2020supervised} in the following sections. Similar to the general contrastive learning like SimCLR, the proposed method contains the following components:
\begin{itemize}

 \item \emph{Data Augmentation} module \cite{chen2020simple,tian2019contrastive,hjelm2018learning}. 
For a batch of data with size $N$, the general operation of contrastive learning is to generate multiview of transformed $t(\mathbf{x})$ with stochastic data augmentations ${t}\sim\mathcal{T}$. In our work, for each input sample, $\mathbf{x}$, we generate two random augmentations $\mathbf{\tilde{x}} = t(\mathbf{x})$ from the original view of the data.

 \item \emph{Encoder Network} $f(\cdot)$ that maps $\mathbf{\tilde{x}}$ to a lower dimension. The encoder is shared by all views. In our case, we work on ResNet-18 and ResNet-50.  

 \item \emph{Projection Network} $g(\cdot)$ that maps $f(\mathbf{\tilde{x}})$ to a latent space $\mathbf{z}=g(f(\mathbf{\tilde{x}}))$. The $g(\cdot)$ could be either a multi-layer perceptron \cite{hastie2001statisticallearning} or just a single linear layer of size. During the downstream linear evaluation, this layer can be removed and replaced with a classification head. 

 \item \emph{exponential mapping} that maps Euclidean vectors to the hyperbolic space.
   
\end{itemize}
Given the proposed HCL framework, for a set of $N$ samples, $\{\mathbf{x}_k\}_{k=1...N}$, in a batch, we augment each data to generate two views. This will result in $2N$ samples in a batch. Let $i\in I\equiv\{1...2N\}$ be the index of an arbitrary augmented instance, and let $j(i)$ be the index of the other samples in the same batch. The self-supervised contrastive learning (e.g., \cite{chen2020simple,tian2019contrastive,hjelm2018learning}) takes the following loss form to pull positive pairs together and push negatives away from the anchor in the latent space. 
\begin{equation}
  \mathcal{L}^{self}
  =\sum_{i\in I}\mathcal{L}_i^{self}
  =-\sum_{i\in I}\log{
  \frac{\text{exp}\left(\mathbf{z}_i\cdot\mathbf{z}_{j(i)}/\tau\right)}{\sum\limits_{a\in A(i)}\text{exp}\left(\mathbf{z}_i\cdot\mathbf{z}_a/\tau\right)}
  }
  \label{eqn:self_loss}
\end{equation} 

 In the above equation, $\mathbf{z}=g(f(\mathbf{\tilde{x}}))$. Usually, $\mathbf{z}$ is normalized before the loss calculation so that features lie on a unit hypersphere. The $\cdot$ symbol denotes the inner (dot) product, $\tau\in\mathcal{R}^+$ is a scalar temperature parameter, and $A(i)\equiv I\setminus\{i\}$. The index $i$ indicates the anchor, index $j(i)$ is its positive pair, and the other $2(N-1)$ indices ($\{k\in A(i)\setminus\{j(i)\}$) indicate the negatives of the anchor. For each anchor $i$, there is $1$ positive pair and $2N - 2$ negative pairs. The denominator has a total of $2N - 1$ terms (the positive and negatives). 

A distance could be defined with the cosine similarity implemented with a squared Euclidean distance between normalized vectors as follow

\begin{equation}
    D_{cos}(\mathbf{z}_i, \mathbf{z}_j)  = \norm{ \frac{\mathbf{z}_i}{\norm{\mathbf{z}_i}_2} - \frac{\mathbf{z}_j}{\norm{\mathbf{z}_j}_2} }^2_2 
 = 2 - 2 \frac{ \mathbf{z}_i \cdot \mathbf{z}_j }{\norm{\mathbf{z}_i}_2 \cdot \norm{\mathbf{z}_j}_2}
\label{eq:dcos}
\end{equation}

In our proposed HCL, the loss function is defined as    
\begin{equation}
  \mathcal{L}^{self}_{hyp}
  =\sum_{i\in I}\mathcal{L}_{hyp_i}^{self}
  =-\sum_{i\in I}\log{
  \frac{\text{exp}\left(-D(\mathbf{z}_i, \mathbf{z}_{j(i)})/\tau\right)}{\sum\limits_{a\in A(i)}\text{exp}\left(-D(\mathbf{z}_i, \mathbf{z}_a)/\tau\right)}
  }
  \label{eqn:hcl}
\end{equation} 
where $D$ is the distance measurement like $D_{cos}$ or $D_{hyp}$. In our case we project $\mathbf{z}$ to the hyperbolic space and use the pre-defined hyperbolic distance $D_{hyp}$ for distance measurement.

\subsection{Supervised Hyperbolic Contrastive Learning}
In self-supervised pretraining, class labels are unknown. For supervised contrastive learning, the contrastive loss was generalized to handle more positives and negatives with information of class labels. For a given dataset $\{\mathbf{x}_k,\mathbf{y}_k\}_{k=1...N}$, the supervised contrastive (SupCon) loss proposed by \cite{khosla2020supervised} is 
\begin{equation}
  \mathcal{L}^{sup}
  =\sum_{i\in I}\mathcal{L}_{i}^{sup}
  =\sum_{i\in I}\frac{-1}{|P(i)|}\sum_{p\in P(i)}\log{\frac{\text{exp}\left(\mathbf{z}_i\cdot\mathbf{z}_p/\tau\right)}{\sum\limits_{a\in A(i)}\text{exp}\left(\mathbf{z}_i\cdot\mathbf{z}_a/\tau\right)}}
  \label{eqn:supervised_loss}
\end{equation}
 where $P(i)\equiv\{p\in A(i):\mathbf{y}_p=\mathbf{y}_i\}$ is the set of indices of all positives in a batch distinct from $i$ (i.e., the augment of $\mathbf{x}_i$ as well as any of the remaining samples with the same label), and $|P(i)|$ is its cardinality. The summation over negatives in the contrastive denominator of Eq. \ref{eqn:self_loss} is also preserved to improve the ability of to discriminate between signal and noise (negatives). 

 Following the construction of HCL, the Supervised Hyperbolic Contrastive loss could be easily constructed as 
\begin{equation}
  \mathcal{L}^{sup}_{hyp}
  =\sum_{i\in I}\frac{-1}{|P(i)|}\sum_{p\in P(i)}\log{\frac{\text{exp}\left(-D(\mathbf{z}_i, \mathbf{z}_p)/\tau\right)}{\sum\limits_{a\in A(i)}\text{exp}\left(-D(\mathbf{z}_i, \mathbf{z}_a)/\tau\right)}}
  \label{eqn:shcl}
\end{equation}
\label{sec:method-sup}

\vspace{-0.5cm}

\subsection{Adversarial Robustness of HCL}



 One of the most popular approaches to mitigate the effect of adversarial perturbation is adversarial training (AT) \cite{madry2017towards}, which trains the neural network with worst-case adversarial examples. Very recently, the connection between self-supervised contrastive learning and adversarial training has been built to develop label-efficient and robust models \cite{jiang2020robust, kim2020adversarial}.   

The work of Robust Contrastive Learning (RoCL) \cite{kim2020adversarial} performs instance-wise adversarial attack with 
\begin{equation}  
    \mathbf{\tilde{x}}^{i+1} = \Pi_{B(\mathbf{\tilde{x}},\epsilon)} (\mathbf{\tilde{x}}^{i} + \alpha \texttt{sign}(\nabla_{\mathbf{\tilde{x}}^{i}}\mathcal{L}^{self}( \mathbf{\tilde{x}},  \mathbf{\tilde{x}}^+,   \{\mathbf{\tilde{x}}^-\})))
    \label{equation:instance}
\end{equation}
where $\mathbf{\tilde{x}}$ is augmented anchor point, $\mathbf{\tilde{x}}^+$ and$\mathbf{\tilde{x}}^-$ are its positive and negative pairs. $B( \mathbf{\tilde{x}},\epsilon)$ is the $\ell_{\infty}$ norm-ball around $ \mathbf{\tilde{x}}$ with radius $\epsilon$, and $\Pi$ is the projection function for norm-ball. To learn robust representation via
self-supervised contrastive learning, the adversarial learning objective for an instance-wise attack following the min-max formulation is
\begin{equation}
    \argmin_{\theta} \mathbb{E}_{(\mathbf{\tilde{x}})\sim\mathbb{D}} [\max_{\delta \in B(\mathbf{\tilde{x}}, \epsilon)} \mathcal{L}^{self}(\mathbf{\tilde{x}}+\delta, \mathbf{\tilde{x}}^+, \{\mathbf{\tilde{x}}^-\}) ]
\end{equation}
where $\theta$ is model parameter and $\mathbb{D}$ is dataset, $\mathbf{\tilde{x}}+\delta$ is the adversarial image $  \mathbf{\tilde{x}}^{adv}$ generated by \emph{instance-wise} attacks (Eq. \ref{equation:instance}). 
 After generating label-free adversarial examples using instance-wise adversarial attacks, the contrastive learning objective Eq. \ref{eqn:self_loss} is used to maximize the similarity between clean examples and their instance-wise perturbation. The final loss of RoCL is a combination of 
$\mathcal{L}^{self}(\mathbf{\tilde{x}}, \{\mathbf{\tilde{x}}^+, \mathbf{\tilde{x}}^{adv}\}, \{\mathbf{\tilde{x}}^-\})$
and $\mathcal{L}^{self} (\mathbf{\tilde{x}}^{adv}, \mathbf{\tilde{x}}^+,\{\mathbf{\tilde{x}}^-\} $ where the first term has extra $\mathbf{\tilde{x}}^{adv}$ as positive pair for anchor $\mathbf{\tilde{x}}$ and the second term uses $\mathbf{\tilde{x}}^{adv}$ as the anchor. 

The RoCL could be easily extended to Robust Hyerbolic Contrastive loss where the output of the network is projected to hyperbolic space for better semantic relationship representation. The objective $ \mathcal{L}^{RHCL}_{hyp}$ is defined as
\begin{equation}\label{equation:RHCL}
\begin{gathered}
    \mathcal{L}^{self}_{hyp}(\mathbf{\tilde{x}}, \{\mathbf{\tilde{x}}^+, \mathbf{\tilde{x}}^{adv}\}, \{\mathbf{\tilde{x}}^-\}) + \lambda \mathcal{L}^{self}_{hyp}(\mathbf{\tilde{x}}^{adv}, \mathbf{\tilde{x}}^+,\{\mathbf{\tilde{x}}^-\})
\end{gathered}
\end{equation}
where $\lambda$ is a regularization parameter.
\label{sec:method-adv}

\section{Experiments and Results}

We conduct comprehensive experiments to cover hyperbolic contrastive learning in three directions: self-supervised domain, supervised domain, and adversarial robustness evaluation. To demonstrate the effectiveness and generality of our method, we verify each proposed method on a variety of datasets. We first introduce the datasets and contrastive methods in Section \ref{exp:data}. The implementation details are given in Section \ref{exp:detail}. We present the experiment results in Section \ref{exp:result} and finally, we show the ablation study results in Section \ref{exp:ablation}.

\subsection{Datasets \& Baseline Approaches}
\label{exp:data}
\emph{Self-supervised Learning} We perform the evaluation of our method on a wide range of datasets include \textbf{CIFAR-10/CIAFR-100} \cite{krizhevsky2009learning}, \textbf{Tiny ImageNet}, and \textbf{ImageNet}~\cite{deng2009imagenet} with different number of classes \cite{tian2020contrastive}. Our proposed HCL is plugged in on the SimCLR. We compare the linear evaluation result with SimCLR~\cite{chen2020simple}.

\emph{Supervised Learning} As an extension of HCL, SHCL uses label information for positive and negative pair distinguishment. We compare our proposed SHCL with SupCon \cite{khosla2020supervised} and cross-entropy by measuring classification accuracy on \textbf{CIFAR-10/CIAFR-100}. 

\emph{Adversarial Robust Learning} We conduct adversarial attack with our proposed RHCL on \textbf{CIFAR-10} and compare the result with SimCLR, HCL, and the recent fully self-supervised robustness learning work RoCL \cite{kim2020adversarial}.

\subsection{Implementation Details}

\emph{Self-supervised Learning} The proposed HCL is aiming at exploring the data representation power of hyperbolic space for contrastive learning. Thus the training components like backbone networks, losses, optimizers, \etc are agnostic. For our method and baseline training, we keep the same training settings when making comparisons. Larger gains are possible with further hyper-parameter tuning.

We follow the code implementation and hyper-parameter of \cite{peng2022crafting}. For small datasets (\ie, CIFAR-10/100 and Tiny ImageNet), we use the same training setup in \textit{all} experiments. At the pretrain stage, we train ResNet-18 \cite{he2016deep} for 200 epochs with a batch size of 512 and a cosine-annealed learning rate of 0.5. The linear classifier is trained for 100 epochs with an initial learning rate of 10.0 multiplied by 0.1 at 60\textit{th} and 80\textit{th} epochs.

For experiments on ImageNet, we use ResNet-50 as the backbone. Since the data size is larger, we train SimCLR and our method with the batch size of 1024 and cosine-annealed learning rate of 0.6 for faster convergence. We adopt the same setting as in \cite{peng2022crafting} for training the linear classifier.

When projecting the data from Euclidian space to hyperbolic space, we define the curvature $c=0.1$ except for CIFAR-10 $c=0.6$. All the experiments are conducted on 4-GPUs. We use SGD optimizer with momentum of 0.9, weight decay of $10^{-4}$ and $0$ for pre-train and linear evaluation, respectively.

\emph{Supervised Learning}
Following the work of \cite{khosla2020supervised}, we experiment with ResNet-50 for supervised classification training. We pretrain 200 epochs of SupCon and our SHCL with the same hyper-parameters we used in self-supervised learning. We then freeze the pretrained model but retrain a classification head using the same hyper-parameters for both models. We tune the curvature $c$ with $0.1, 0.2$. We also tune the baseline cross entropy case a little bit so that it could reach a comparable result on both CIFAR-10 and CIFAR-100. 

\emph{Adversarial Robust Learning}
We use the code structure and evaluation procedure by the first work that explored the adversarial robustness of self-supervised contrastive learning RoCL \cite{kim2020adversarial}. Since our purpose is to explore the robustness of the proposed RHCL, we pretrain all the methods for 200 epochs with the same hyper-parameters used in self-supervised learning. Similar to RoCL, we use ResNet-18 trained on CIFAR-10. For all baselines and our method, we train with $\ell_\infty$ attacks with the same attack strength of $\epsilon = 8/255$. We perform two kinds of evaluations. 1) Since general self-supervised learning involves two steps, pretraining and linear evaluation. After training our proposed RHCL with adversarial perturbations, we perform linear evaluation with adversarial examples after fixing the pretrained backbone. The linear evaluation for \textit{all} experiments were trained 150 epochs following the RoCL paper, all the other hyper-parameters are exactly the same as the literature \cite{kim2020adversarial}. 2) The whole pretrained network could be finetuned with adversarial examples. We perform supervised adversarial finetune \cite{madry2017towards}. Same as linear evaluation, the finetune also trained the network for 150 epochs. All other parameters including $\lambda$ are exactly the same as in the work of \cite{kim2020adversarial} for evaluation. 

\label{exp:detail}

\subsection{Results}
\label{exp:result}
\bfsection{Self-supervised Learning Results}

In this section, we verify our method with linear probe following the common self-supervised contrastive learning procedure. We freeze pre-trained weights of the encoder and train a supervised linear classifier on top of it. We then report the Top-1 classification accuracy on the validation set.

Our results on CIFAR-10/CIFAR-100, Tiny ImageNet (Tiny IN), ImageNet (IN-1K) and subset of ImageNet with 100/200 classes (IN-100, IN-200) are shown in Tab. \ref{tab:IN}. With the exact same training process for \textit{all} experiments, \textit{HCL} consistently improves baseline methods SimCLR by at least $0.44$ (IN-100). For CIFAR-100, the gain is $6.98$ without any heavy parameter tuning. For the largest dataset IN-1K, the linear classification improvement is $1.12$ with our method. Note that we did not make much effort to tune the hyper-parameter including the curvature $c$. In the ablation study we show that when $c=0.6$ we got the best result on CIFAR-10 compared with other $c$ values under the same condition. We report the CIFAR-10 best result after searching $c$ in Tab. \ref{tab:IN}. For all other datasets, we use default $c=0.1$ without searching. It is shown that the proposed HCL is an effective and easy to plug-in method for general contrastive learning and does not require heavy parameter tuning.

\begin{table}[ht]
    \centering
    \small
    \begin{tabular}{l|c|c|c|c}
        \toprule
        \multirow{1}{*}{\bf Data} & 
        \multirow{1}{*}{\bf Arch.} &
        \multirow{1}{*}{\bf Epoch} & \bf SimCLR & \bf HCL \\
        \midrule
        CIFAR-10 & ResNet-18 & 200 & 86.58 & \textbf{87.98} \\
      
        \hline
        CIFAR-100 & ResNet-18 & 200& 48.91 & \textbf{55.89} \\
       
        \hline
        TinyIN & ResNet-18 & 200 & 44.36 & \textbf{45.02} \\
       
        \hline
        IN-100 & ResNet-50 & 100 & 75.26 & \textbf{75.70} \\
        
        \hline
        IN-200 & ResNet-50 & 100 & 73.18 & \textbf{73.64}  \\     

        \hline
        IN-1K & ResNet-50 & 100 & 57.81 & \textbf{58.93}  \\ 
        \bottomrule
    \end{tabular}
    \caption{Comparison of \textit{SimCLR} and our \textit{HCL} with linear classification results on different benchmarks. Models are pre-trained with the same training setup within a dataset for a fair comparison.}
    \label{tab:IN}
    \vspace{-0.2cm}
\end{table}

\bfsection{Supervised Classification}

We plug in our method of projecting the embeddings to hyperbolic space and calculate the proposed supervised objective in the new space.  
Table \ref{tab:sup} shows that our SHCL could generalize better than SupCon \cite{khosla2020supervised} and cross-entropy (CE) on CIFAR-10 and CIFAR-100 when training on ResNet-50. Both SupCon and SHCL are trained with the same hyper-parameter. For CE, we tune the model a little bit and trained 300 epochs in total for CIFAR-10 so it can reach a comparable result with others. Our SHCL could reach $95.1$ accuracy with 200 epochs pertaining on CIAR-10 compared with SupCon. Both experiments show that our method is slightly superior to SupCon in the supervised classification domain.  

\begin{table}[ht]
    \centering
    \small
    \begin{tabular}{l|c|c|c|c}
        \toprule
        \multirow{1}{*}{\bf Data} & 
        \multirow{1}{*}{\bf Arch.} &
        \multirow{1}{*}{\bf CE} & \bf SupCon & \bf SHCL \\
        \midrule
        CIFAR-10 & ResNet-50 & 93.90 & 94.84 & \textbf{95.1} \\
      
        \hline
        CIFAR-100 & ResNet-50 & 75.38 & 83.01 & \textbf{83.58} \\
       
        \bottomrule
    \end{tabular}
    \caption{Top-1 classification accuracy on ResNet-50 \cite{he2016deep} for various datasets. We compare \textit{cross-entropy} training, \textit{SupCon} \cite{khosla2020supervised}, and our \textit{SHCL}. Baseline numbers in the table are based on our re-implementation.}
    \label{tab:sup}
\end{table}

\bfsection{Adversarial Robustness}

Compared with standard training, adversarial training is computationally more expensive. The main purpose of this work is to explore the superior embedding ability of hyperbolic space. As an extension study of the proposed HCL, in this section we focus on examining the robustness of our method to adversarial attack rather than pursuing the state-of-the-art adversarial robustness. Thus we pretrain the model for 200 epochs. We first compare RHCL against SimCLR\cite{chen2020simple} and our proposed HCL. Tab. \ref{tab:adv} shows experimental results with white box attacks on CIFAR-10. $A_{nat}$ is the accuracy of clean image and the $\ell_{\infty}$ is the adversarial attack.

The result indicates that SimCLR is extremely vulnerable to adversarial attacks while HCL shows slight robustness to the adversarial attack. HCL has the best performance on clean data evaluation. Both SimCLR and HCL are actually very vulnerable to adversarial attacks. However, RHCL achieves high robust accuracy (29.94) against the target $\ell_\infty$ attacks compared with RoCL (26.57). Both RoCL and RHCL show impressive improvements regarding the adversarial attack. Note these numbers are gained without using any label data. Our RHCL performs better than other self-supervised methods in the table regarding both clean and adversarial samples. Moreover, when the model is finetuned with adversarial training by \cite{madry2017towards}, the performance is further boosted.

\begin{table}[t]
    \centering
    \small
{
  \begin{tabular}{cccc} %
    \toprule
    \multirow{2}{*}{\makecell{\\Train type}}&\multirow{2}{*}{\makecell{\\Method}}&\multicolumn{2}{c}{CIFAR-10}\\ 
    \cmidrule(r){3-4}
    {}&{}&$A_{nat}$&$\ell_{\infty}$\\
    \midrule \midrule 
    \multirow{4}{*}{\makecell{Self-supervised}} & SimCLR &79.71&0.0\\
    {}&\textbf{HCL}&\textbf{83.02} &0.09 \\
   
    {}&RoCL& 72.94& 26.57\\

    {}&\textbf{RHCL}& 73.45 & \textbf{29.94}\\
    
    \midrule
    \multirow{1}{*}{\makecell{Self-supervised+finetune}}
    {}&\textbf{RHCL+AT}&77.9&\textbf{32.61}\\

    \bottomrule
  \end{tabular}
  }
    \caption{ Experimental results with white box attacks on ResNet-18 trained on the CIFAR-10. AT denotes the supervised adversarial training\cite{madry2017towards}. \textit{RoCL} is a baseline model from \cite{kim2020adversarial}. $A_{nat}$ is the accuracy of clean images. All models are trained with $\ell_{\infty}$.}
      \label{tab:adv}

\end{table}

\subsection{Ablation Studies}
\label{exp:ablation}

\begin{figure}
    \centering
        \vspace{-3mm}
    \includegraphics[width=0.9\linewidth]{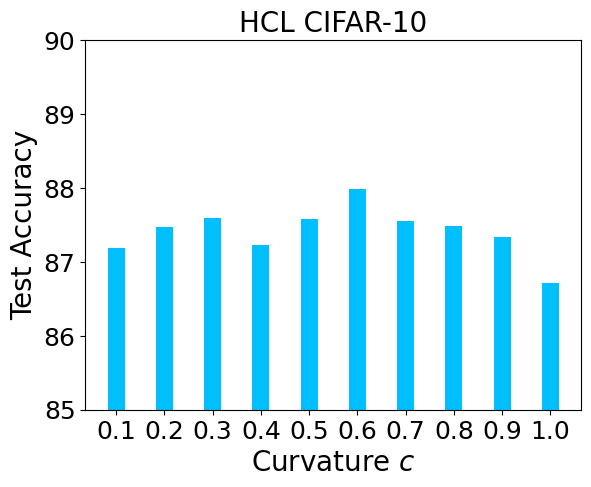}
    \vspace{-2mm}
    \caption{Comparison of CIFAR-10 with \textit{HCL} with different curvature $c$.}
    \label{fig:curve}
\end{figure}

\bfsection{Curvature}
We experimented the effect of curvature $c$ in HCL on CIFAR-10 and present the comparison result in Fig. \ref{fig:curve} as well as Tab. \ref{tab:ab1}. We evaluate the linear accuracy with $c$ changes from $0.1$ to $1$. It can be seen that our method is slightly affected by $c$. When $c$ reaches 0.6, the model has the best accuracy and thus the best representation power. 

\begin{table}[ht]
    \centering
    \small
    \begin{tabular}{l |c|c|c|c|c}
        \toprule
        Curvature& 0.1 & 0.2 & 0.3 & 0.4 & 0.5 \\
        \midrule
        Accuracy & 87.19 & 87.47 & 87.59& 87.23& 87.58 \\
        \midrule
         Curvature& \bf 0.6 & 0.7 & 0.8 & 0.9 & 1.0 \\
        \midrule
          Accuracy &\bf 87.98& 87.55& 87.49& 87.33& 86.72 \\
        \bottomrule
    \end{tabular}
    \caption{Classification accuracy of \textit{HCL} trained on CIFAR-10 with different curvature $c$}
    \label{tab:ab1}
\end{table}

\bfsection{Normalization}
Usually self-supervised learning involves normalizing the output of the network before calculating the loss. In our proposed method we need to project the output to hyperbolic space. We found whether normalizing the embedding before mapping to the \Poincare affects the model performance. We evaluate HCL on CIFAR-10 and CIFAR-100 with and without normalization. Our experiment shows that the accuracy of CIFAR-10 drops from $87.98$ to $85.37$ without normalization. The performance of CIFAR-100 drops from $55.89$ to $46.44$. The normalization layer is vital in our model. A similar scenario has been observed in \cite{khosla2020supervised}.

\section{Discussion}

The hyperbolic space owns better tree-structure representation power than Euclidean space. Inspired by the recent success of hyperbolic embedding in NLP and other computer vision domains, in this paper, we explore contrastive learning in hyperbolic space. We propose a new contrastive learning framework for self-supervised image representation learning. The proposed method is evaluated on different small to large-scale datasets and shows promising results. We further extend hyperbolic contrastive learning to supervised contrastive learning and demonstrate its superior performance on different classification tasks. Self-supervised representation learning usually involves learning a pretrained model so that downstream tasks could be fine-tuned faster or gain higher accuracy. Lately, the research attempt has been made to combine adversarial training with self-supervision for robust pretrained models that can be rapidly used by downstream tasks. We explore the adversarial robustness of proposed hyperbolic contrastive learning. To the best of our knowledge, this is the first work that attempts to build contrastive models in hyperbolic space and explore the self-supervised adversarial robustness in this space. In our study, we show some preliminary results in this direction to give some insights for future studies. There are some other questions that have not been explored such as whether the hyperbolic space representation will benefit other downstream tasks. Could we explicitly guide the model with hierarchy information when labels are available? We leave these questions for future studies.  

{\small
\bibliographystyle{ieee_fullname}
\bibliography{iclr2023_conference}

\begin{thebibliography}{10}\itemsep=-1pt

\bibitem{alayrac2019labels}
Jean-Baptiste Alayrac, Jonathan Uesato, Po-Sen Huang, Alhussein Fawzi, Robert
  Stanforth, and Pushmeet Kohli.
\newblock Are labels required for improving adversarial robustness?
\newblock {\em Advances in Neural Information Processing Systems}, 32, 2019.

\bibitem{arora2019theoretical}
Sanjeev Arora, Hrishikesh Khandeparkar, Mikhail Khodak, Orestis Plevrakis, and
  Nikunj Saunshi.
\newblock A theoretical analysis of contrastive unsupervised representation
  learning.
\newblock {\em arXiv preprint arXiv:1902.09229}, 2019.

\bibitem{atigh2022hyperbolic}
Mina~Ghadimi Atigh, Julian Schoep, Erman Acar, Nanne van Noord, and Pascal
  Mettes.
\newblock Hyperbolic image segmentation.
\newblock In {\em Proceedings of the IEEE/CVF Conference on Computer Vision and
  Pattern Recognition}, pages 4453--4462, 2022.

\bibitem{bachman2019learning}
Philip Bachman, R~Devon Hjelm, and William Buchwalter.
\newblock Learning representations by maximizing mutual information across
  views.
\newblock {\em Advances in neural information processing systems}, 32, 2019.

\bibitem{boudiaf2020unifying}
Malik Boudiaf, J{\'e}r{\^o}me Rony, Imtiaz~Masud Ziko, Eric Granger, Marco
  Pedersoli, Pablo Piantanida, and Ismail~Ben Ayed.
\newblock A unifying mutual information view of metric learning: cross-entropy
  vs. pairwise losses.
\newblock In {\em European conference on computer vision}, pages 548--564.
  Springer, 2020.

\bibitem{bronstein2017geometric}
Michael~M Bronstein, Joan Bruna, Yann LeCun, Arthur Szlam, and Pierre
  Vandergheynst.
\newblock Geometric deep learning: going beyond euclidean data.
\newblock {\em IEEE Signal Processing Magazine}, 34(4):18--42, 2017.

\bibitem{cannon1997hyperbolic}
James~W Cannon, William~J Floyd, Richard Kenyon, Walter~R Parry, et~al.
\newblock Hyperbolic geometry.
\newblock {\em Flavors of geometry}, 31(59-115):2, 1997.

\bibitem{carlini2017towards}
Nicholas Carlini and David Wagner.
\newblock Towards evaluating the robustness of neural networks.
\newblock In {\em 2017 ieee symposium on security and privacy (sp)}, pages
  39--57. Ieee, 2017.

\bibitem{carlucci2019domain}
Fabio~M Carlucci, Antonio D'Innocente, Silvia Bucci, Barbara Caputo, and
  Tatiana Tommasi.
\newblock Domain generalization by solving jigsaw puzzles.
\newblock In {\em Proceedings of the IEEE/CVF Conference on Computer Vision and
  Pattern Recognition}, pages 2229--2238, 2019.

\bibitem{carmon2019unlabeled}
Yair Carmon, Aditi Raghunathan, Ludwig Schmidt, John~C Duchi, and Percy~S
  Liang.
\newblock Unlabeled data improves adversarial robustness.
\newblock {\em Advances in Neural Information Processing Systems}, 32, 2019.

\bibitem{caron2020unsupervised}
Mathilde Caron, Ishan Misra, Julien Mairal, Priya Goyal, Piotr Bojanowski, and
  Armand Joulin.
\newblock Unsupervised learning of visual features by contrasting cluster
  assignments.
\newblock {\em Advances in Neural Information Processing Systems},
  33:9912--9924, 2020.

\bibitem{chen2020simple}
Ting Chen, Simon Kornblith, Mohammad Norouzi, and Geoffrey Hinton.
\newblock A simple framework for contrastive learning of visual
  representations.
\newblock In {\em International conference on machine learning}, pages
  1597--1607. PMLR, 2020.

\bibitem{chen2020big}
Ting Chen, Simon Kornblith, Kevin Swersky, Mohammad Norouzi, and Geoffrey~E
  Hinton.
\newblock Big self-supervised models are strong semi-supervised learners.
\newblock {\em Advances in neural information processing systems},
  33:22243--22255, 2020.

\bibitem{chen2020adversarial}
Tianlong Chen, Sijia Liu, Shiyu Chang, Yu Cheng, Lisa Amini, and Zhangyang
  Wang.
\newblock Adversarial robustness: From self-supervised pre-training to
  fine-tuning.
\newblock In {\em Proceedings of the IEEE/CVF Conference on Computer Vision and
  Pattern Recognition}, pages 699--708, 2020.

\bibitem{chen2021intriguing}
Ting Chen, Calvin Luo, and Lala Li.
\newblock Intriguing properties of contrastive losses.
\newblock {\em Advances in Neural Information Processing Systems}, 34, 2021.

\bibitem{chen2020improved}
Xinlei Chen, Haoqi Fan, Ross Girshick, and Kaiming He.
\newblock Improved baselines with momentum contrastive learning.
\newblock {\em arXiv preprint arXiv:2003.04297}, 2020.

\bibitem{chopra2005learning}
Sumit Chopra, Raia Hadsell, and Yann LeCun.
\newblock Learning a similarity metric discriminatively, with application to
  face verification.
\newblock In {\em 2005 IEEE Computer Society Conference on Computer Vision and
  Pattern Recognition (CVPR'05)}, volume~1, pages 539--546. IEEE, 2005.

\bibitem{deng2009large}
Jia Deng.
\newblock A large-scale hierarchical image database.
\newblock {\em Proc. of IEEE Computer Vision and Pattern Recognition, 2009},
  2009.

\bibitem{deng2009imagenet}
Jia Deng, Wei Dong, Richard Socher, Li-Jia Li, Kai Li, and Li Fei-Fei.
\newblock Imagenet: A large-scale hierarchical image database.
\newblock In {\em CVPR}, 2009.

\bibitem{ermolov2022hyperbolic}
Aleksandr Ermolov, Leyla Mirvakhabova, Valentin Khrulkov, Nicu Sebe, and Ivan
  Oseledets.
\newblock Hyperbolic vision transformers: Combining improvements in metric
  learning.
\newblock In {\em Proceedings of the IEEE/CVF Conference on Computer Vision and
  Pattern Recognition}, pages 7409--7419, 2022.

\bibitem{everingham2010pascal}
Mark Everingham, Luc Van~Gool, Christopher~KI Williams, John Winn, and Andrew
  Zisserman.
\newblock The pascal visual object classes (voc) challenge.
\newblock {\em International journal of computer vision}, 88(2):303--338, 2010.

\bibitem{eykholt2018robust}
Kevin Eykholt, Ivan Evtimov, Earlence Fernandes, Bo Li, Amir Rahmati, Chaowei
  Xiao, Atul Prakash, Tadayoshi Kohno, and Dawn Song.
\newblock Robust physical-world attacks on deep learning visual classification.
\newblock In {\em Proceedings of the IEEE conference on computer vision and
  pattern recognition}, pages 1625--1634, 2018.

\bibitem{fang2021kernel}
Pengfei Fang, Mehrtash Harandi, and Lars Petersson.
\newblock Kernel methods in hyperbolic spaces.
\newblock In {\em Proceedings of the IEEE/CVF International Conference on
  Computer Vision}, pages 10665--10674, 2021.

\bibitem{faster2015towards}
RCNN Faster.
\newblock Towards real-time object detection with region proposal networks.
\newblock {\em Advances in neural information processing systems},
  9199(10.5555):2969239--2969250, 2015.

\bibitem{fukushima1982neocognitron}
Kunihiko Fukushima and Sei Miyake.
\newblock Neocognitron: A self-organizing neural network model for a mechanism
  of visual pattern recognition.
\newblock In {\em Competition and cooperation in neural nets}, pages 267--285.
  Springer, 1982.

\bibitem{ganea2018hyperbolic}
Octavian Ganea, Gary B{\'e}cigneul, and Thomas Hofmann.
\newblock Hyperbolic neural networks.
\newblock {\em Advances in neural information processing systems}, 31, 2018.

\bibitem{gao2021curvature}
Zhi Gao, Yuwei Wu, Yunde Jia, and Mehrtash Harandi.
\newblock Curvature generation in curved spaces for few-shot learning.
\newblock In {\em Proceedings of the IEEE/CVF International Conference on
  Computer Vision}, pages 8691--8700, 2021.

\bibitem{gidaris2018unsupervised}
Spyros Gidaris, Praveer Singh, and Nikos Komodakis.
\newblock Unsupervised representation learning by predicting image rotations.
\newblock {\em arXiv preprint arXiv:1803.07728}, 2018.

\bibitem{girshick2014rich}
Ross Girshick, Jeff Donahue, Trevor Darrell, and Jitendra Malik.
\newblock Rich feature hierarchies for accurate object detection and semantic
  segmentation.
\newblock In {\em Proceedings of the IEEE conference on computer vision and
  pattern recognition}, pages 580--587, 2014.

\bibitem{goldberger2004neighbourhood}
Jacob Goldberger, Geoffrey~E Hinton, Sam Roweis, and Russ~R Salakhutdinov.
\newblock Neighbourhood components analysis.
\newblock {\em Advances in neural information processing systems}, 17, 2004.

\bibitem{goodfellow2014explaining}
Ian~J Goodfellow, Jonathon Shlens, and Christian Szegedy.
\newblock Explaining and harnessing adversarial examples.
\newblock {\em arXiv preprint arXiv:1412.6572}, 2014.

\bibitem{guler2018densepose}
R{\i}za~Alp G{\"u}ler, Natalia Neverova, and Iasonas Kokkinos.
\newblock Densepose: Dense human pose estimation in the wild.
\newblock In {\em Proceedings of the IEEE conference on computer vision and
  pattern recognition}, pages 7297--7306, 2018.

\bibitem{guo2022clipped}
Yunhui Guo, Xudong Wang, Yubei Chen, and Stella~X Yu.
\newblock Clipped hyperbolic classifiers are super-hyperbolic classifiers.
\newblock In {\em Proceedings of the IEEE/CVF Conference on Computer Vision and
  Pattern Recognition}, pages 11--20, 2022.

\bibitem{guo2022hcsc}
Yuanfan Guo, Minghao Xu, Jiawen Li, Bingbing Ni, Xuanyu Zhu, Zhenbang Sun, and
  Yi Xu.
\newblock Hcsc: Hierarchical contrastive selective coding.
\newblock In {\em Proceedings of the IEEE/CVF Conference on Computer Vision and
  Pattern Recognition}, pages 9706--9715, 2022.

\bibitem{gupta2022higher}
Rohit Gupta, Naveed Akhtar, Ajmal Mian, and Mubarak Shah.
\newblock On higher adversarial susceptibility of contrastive self-supervised
  learning.
\newblock {\em arXiv preprint arXiv:2207.10862}, 2022.

\bibitem{hadsell2006dimensionality}
Raia Hadsell, Sumit Chopra, and Yann LeCun.
\newblock Dimensionality reduction by learning an invariant mapping.
\newblock In {\em 2006 IEEE Computer Society Conference on Computer Vision and
  Pattern Recognition (CVPR'06)}, volume~2, pages 1735--1742. IEEE, 2006.

\bibitem{hastie2001statisticallearning}
Trevor Hastie, Robert Tibshirani, and Jerome Friedman.
\newblock {\em The Elements of Statistical Learning}.
\newblock Springer Series in Statistics. Springer New York Inc., New York, NY,
  USA, 2001.

\bibitem{he2020momentum}
Kaiming He, Haoqi Fan, Yuxin Wu, Saining Xie, and Ross Girshick.
\newblock Momentum contrast for unsupervised visual representation learning.
\newblock In {\em Proceedings of the IEEE/CVF conference on computer vision and
  pattern recognition}, pages 9729--9738, 2020.

\bibitem{he2017mask}
Kaiming He, Georgia Gkioxari, Piotr Doll{\'a}r, and Ross Girshick.
\newblock Mask r-cnn.
\newblock In {\em Proceedings of the IEEE international conference on computer
  vision}, pages 2961--2969, 2017.

\bibitem{he2016deep}
Kaiming He, Xiangyu Zhang, Shaoqing Ren, and Jian Sun.
\newblock Deep residual learning for image recognition.
\newblock In {\em Proceedings of the IEEE conference on computer vision and
  pattern recognition}, pages 770--778, 2016.

\bibitem{hinton2006fast}
Geoffrey~E Hinton, Simon Osindero, and Yee-Whye Teh.
\newblock A fast learning algorithm for deep belief nets.
\newblock {\em Neural computation}, 18(7):1527--1554, 2006.

\bibitem{hjelm2018learning}
R~Devon Hjelm, Alex Fedorov, Samuel Lavoie-Marchildon, Karan Grewal, Phil
  Bachman, Adam Trischler, and Yoshua Bengio.
\newblock Learning deep representations by mutual information estimation and
  maximization.
\newblock {\em arXiv preprint arXiv:1808.06670}, 2018.

\bibitem{hu2017learning}
Weihua Hu, Takeru Miyato, Seiya Tokui, Eiichi Matsumoto, and Masashi Sugiyama.
\newblock Learning discrete representations via information maximizing
  self-augmented training.
\newblock In {\em International conference on machine learning}, pages
  1558--1567. PMLR, 2017.

\bibitem{jaiswal2020survey}
Ashish Jaiswal, Ashwin~Ramesh Babu, Mohammad~Zaki Zadeh, Debapriya Banerjee,
  and Fillia Makedon.
\newblock A survey on contrastive self-supervised learning.
\newblock {\em Technologies}, 9(1):2, 2020.

\bibitem{jiang2020robust}
Ziyu Jiang, Tianlong Chen, Ting Chen, and Zhangyang Wang.
\newblock Robust pre-training by adversarial contrastive learning.
\newblock {\em Advances in Neural Information Processing Systems},
  33:16199--16210, 2020.

\bibitem{kayhan2020translation}
Osman~Semih Kayhan and Jan C~van Gemert.
\newblock On translation invariance in cnns: Convolutional layers can exploit
  absolute spatial location.
\newblock In {\em Proceedings of the IEEE/CVF Conference on Computer Vision and
  Pattern Recognition}, pages 14274--14285, 2020.

\bibitem{khosla2020supervised}
Prannay Khosla, Piotr Teterwak, Chen Wang, Aaron Sarna, Yonglong Tian, Phillip
  Isola, Aaron Maschinot, Ce Liu, and Dilip Krishnan.
\newblock Supervised contrastive learning.
\newblock {\em Advances in Neural Information Processing Systems},
  33:18661--18673, 2020.

\bibitem{khrulkov2020hyperbolic}
Valentin Khrulkov, Leyla Mirvakhabova, Evgeniya Ustinova, Ivan Oseledets, and
  Victor Lempitsky.
\newblock Hyperbolic image embeddings.
\newblock In {\em Proceedings of the IEEE/CVF Conference on Computer Vision and
  Pattern Recognition}, pages 6418--6428, 2020.

\bibitem{kim2020adversarial}
Minseon Kim, Jihoon Tack, and Sung~Ju Hwang.
\newblock Adversarial self-supervised contrastive learning.
\newblock {\em Advances in Neural Information Processing Systems},
  33:2983--2994, 2020.

\bibitem{krizhevsky2009learning}
Alex Krizhevsky, Geoffrey Hinton, et~al.
\newblock Learning multiple layers of features from tiny images.
\newblock 2009.

\bibitem{li2020prototypical}
Junnan Li, Pan Zhou, Caiming Xiong, and Steven~CH Hoi.
\newblock Prototypical contrastive learning of unsupervised representations.
\newblock {\em arXiv preprint arXiv:2005.04966}, 2020.

\bibitem{lin2014microsoft}
Tsung-Yi Lin, Michael Maire, Serge Belongie, James Hays, Pietro Perona, Deva
  Ramanan, Piotr Doll{\'a}r, and C~Lawrence Zitnick.
\newblock Microsoft coco: Common objects in context.
\newblock In {\em European conference on computer vision}, pages 740--755.
  Springer, 2014.

\bibitem{liu2020hyperbolic}
Shaoteng Liu, Jingjing Chen, Liangming Pan, Chong-Wah Ngo, Tat-Seng Chua, and
  Yu-Gang Jiang.
\newblock Hyperbolic visual embedding learning for zero-shot recognition.
\newblock In {\em Proceedings of the IEEE/CVF conference on computer vision and
  pattern recognition}, pages 9273--9281, 2020.

\bibitem{long2015fully}
Jonathan Long, Evan Shelhamer, and Trevor Darrell.
\newblock Fully convolutional networks for semantic segmentation.
\newblock In {\em Proceedings of the IEEE conference on computer vision and
  pattern recognition}, pages 3431--3440, 2015.

\bibitem{madry2017towards}
Aleksander Madry, Aleksandar Makelov, Ludwig Schmidt, Dimitris Tsipras, and
  Adrian Vladu.
\newblock Towards deep learning models resistant to adversarial attacks.
\newblock {\em arXiv preprint arXiv:1706.06083}, 2017.

\bibitem{misra2020self}
Ishan Misra and Laurens van~der Maaten.
\newblock Self-supervised learning of pretext-invariant representations.
\newblock In {\em Proceedings of the IEEE/CVF Conference on Computer Vision and
  Pattern Recognition}, pages 6707--6717, 2020.

\bibitem{montanaro2022rethinking}
Antonio Montanaro, Diego Valsesia, and Enrico Magli.
\newblock Rethinking the compositionality of point clouds through
  regularization in the hyperbolic space.
\newblock {\em arXiv preprint arXiv:2209.10318}, 2022.

\bibitem{moosavi2017universal}
Seyed-Mohsen Moosavi-Dezfooli, Alhussein Fawzi, Omar Fawzi, and Pascal
  Frossard.
\newblock Universal adversarial perturbations.
\newblock In {\em Proceedings of the IEEE conference on computer vision and
  pattern recognition}, pages 1765--1773, 2017.

\bibitem{musgrave2020metric}
Kevin Musgrave, Serge Belongie, and Ser-Nam Lim.
\newblock A metric learning reality check.
\newblock In {\em European Conference on Computer Vision}, pages 681--699.
  Springer, 2020.

\bibitem{nickel2017poincare}
Maximillian Nickel and Douwe Kiela.
\newblock Poincar{\'e} embeddings for learning hierarchical representations.
\newblock {\em Advances in neural information processing systems}, 30, 2017.

\bibitem{nickel2018learning}
Maximillian Nickel and Douwe Kiela.
\newblock Learning continuous hierarchies in the lorentz model of hyperbolic
  geometry.
\newblock In {\em International Conference on Machine Learning}, pages
  3779--3788. PMLR, 2018.

\bibitem{noroozi2016unsupervised}
Mehdi Noroozi and Paolo Favaro.
\newblock Unsupervised learning of visual representations by solving jigsaw
  puzzles.
\newblock In {\em European conference on computer vision}, pages 69--84.
  Springer, 2016.

\bibitem{oh2016deep}
Hyun Oh~Song, Yu Xiang, Stefanie Jegelka, and Silvio Savarese.
\newblock Deep metric learning via lifted structured feature embedding.
\newblock In {\em Proceedings of the IEEE conference on computer vision and
  pattern recognition}, pages 4004--4012, 2016.

\bibitem{oord2018representation}
Aaron van~den Oord, Yazhe Li, and Oriol Vinyals.
\newblock Representation learning with contrastive predictive coding.
\newblock {\em arXiv preprint arXiv:1807.03748}, 2018.

\bibitem{oquab2014learning}
Maxime Oquab, Leon Bottou, Ivan Laptev, and Josef Sivic.
\newblock Learning and transferring mid-level image representations using
  convolutional neural networks.
\newblock In {\em Proceedings of the IEEE conference on computer vision and
  pattern recognition}, pages 1717--1724, 2014.

\bibitem{papernot2016limitations}
Nicolas Papernot, Patrick McDaniel, Somesh Jha, Matt Fredrikson, Z~Berkay
  Celik, and Ananthram Swami.
\newblock The limitations of deep learning in adversarial settings.
\newblock In {\em 2016 IEEE European symposium on security and privacy
  (EuroS\&P)}, pages 372--387. IEEE, 2016.

\bibitem{peng2021hyperbolic}
Wei Peng, Tuomas Varanka, Abdelrahman Mostafa, Henglin Shi, and Guoying Zhao.
\newblock Hyperbolic deep neural networks: A survey.
\newblock {\em arXiv preprint arXiv:2101.04562}, 2021.

\bibitem{peng2022crafting}
Xiangyu Peng, Kai Wang, Zheng Zhu, Mang Wang, and Yang You.
\newblock Crafting better contrastive views for siamese representation
  learning.
\newblock In {\em Proceedings of the IEEE/CVF Conference on Computer Vision and
  Pattern Recognition}, pages 16031--16040, 2022.

\bibitem{qi2022adversarial}
Guo-Jun Qi and Mubarak Shah.
\newblock Adversarial pretraining of self-supervised deep networks: Past,
  present and future.
\newblock {\em arXiv preprint arXiv:2210.13463}, 2022.

\bibitem{qin2019adversarial}
Chongli Qin, James Martens, Sven Gowal, Dilip Krishnan, Krishnamurthy
  Dvijotham, Alhussein Fawzi, Soham De, Robert Stanforth, and Pushmeet Kohli.
\newblock Adversarial robustness through local linearization.
\newblock {\em Advances in Neural Information Processing Systems}, 32, 2019.

\bibitem{sarkar2012low}
Rik Sarkar.
\newblock Low distortion delaunay embedding of trees in hyperbolic plane.
\newblock In {\em International symposium on graph drawing}, pages 355--366.
  Springer, 2012.

\bibitem{shimizu2020hyperbolic}
Ryohei Shimizu, Yusuke Mukuta, and Tatsuya Harada.
\newblock Hyperbolic neural networks++.
\newblock {\em arXiv preprint arXiv:2006.08210}, 2020.

\bibitem{simonyan2014very}
Karen Simonyan and Andrew Zisserman.
\newblock Very deep convolutional networks for large-scale image recognition.
\newblock {\em arXiv preprint arXiv:1409.1556}, 2014.

\bibitem{sohn2016improved}
Kihyuk Sohn.
\newblock Improved deep metric learning with multi-class n-pair loss objective.
\newblock {\em Advances in neural information processing systems}, 29, 2016.

\bibitem{szegedy2013intriguing}
Christian Szegedy, Wojciech Zaremba, Ilya Sutskever, Joan Bruna, Dumitru Erhan,
  Ian Goodfellow, and Rob Fergus.
\newblock Intriguing properties of neural networks.
\newblock {\em arXiv preprint arXiv:1312.6199}, 2013.

\bibitem{tian2019contrastive}
Yonglong Tian, Dilip Krishnan, and Phillip Isola.
\newblock Contrastive multiview coding.
\newblock {\em arXiv preprint arXiv:1906.05849}, 2019.

\bibitem{tian2020contrastive}
Yonglong Tian, Dilip Krishnan, and Phillip Isola.
\newblock Contrastive multiview coding.
\newblock In {\em European conference on computer vision}, pages 776--794.
  Springer, 2020.

\bibitem{trinh2019selfie}
Trieu~H Trinh, Minh-Thang Luong, and Quoc~V Le.
\newblock Selfie: Self-supervised pretraining for image embedding.
\newblock {\em arXiv preprint arXiv:1906.02940}, 2019.

\bibitem{ungar2008analytic}
Abraham~Albert Ungar.
\newblock {\em Analytic hyperbolic geometry and Albert Einstein’s special
  theory of relativity}.
\newblock World Scientific, 2008.

\bibitem{ungar2008gyrovector}
Abraham~Albert Ungar.
\newblock A gyrovector space approach to hyperbolic geometry.
\newblock {\em Synthesis Lectures on Mathematics and Statistics}, 1(1):1--194,
  2008.

\bibitem{wang2021understanding}
Feng Wang and Huaping Liu.
\newblock Understanding the behaviour of contrastive loss.
\newblock In {\em Proceedings of the IEEE/CVF conference on computer vision and
  pattern recognition}, pages 2495--2504, 2021.

\bibitem{wang2020understanding}
Tongzhou Wang and Phillip Isola.
\newblock Understanding contrastive representation learning through alignment
  and uniformity on the hypersphere.
\newblock In {\em International Conference on Machine Learning}, pages
  9929--9939. PMLR, 2020.

\bibitem{weinberger2009distance}
Kilian~Q Weinberger and Lawrence~K Saul.
\newblock Distance metric learning for large margin nearest neighbor
  classification.
\newblock {\em Journal of machine learning research}, 10(2), 2009.

\bibitem{weng2021unsupervised}
Zhenzhen Weng, Mehmet~Giray Ogut, Shai Limonchik, and Serena Yeung.
\newblock Unsupervised discovery of the long-tail in instance segmentation
  using hierarchical self-supervision.
\newblock In {\em Proceedings of the IEEE/CVF Conference on Computer Vision and
  Pattern Recognition}, pages 2603--2612, 2021.

\bibitem{wiskott2002slow}
Laurenz Wiskott and Terrence~J Sejnowski.
\newblock Slow feature analysis: Unsupervised learning of invariances.
\newblock {\em Neural computation}, 14(4):715--770, 2002.

\bibitem{wu2018unsupervised}
Zhirong Wu, Yuanjun Xiong, Stella~X Yu, and Dahua Lin.
\newblock Unsupervised feature learning via non-parametric instance
  discrimination.
\newblock In {\em Proceedings of the IEEE conference on computer vision and
  pattern recognition}, pages 3733--3742, 2018.

\bibitem{xie2020unsupervised}
Qizhe Xie, Zihang Dai, Eduard Hovy, Thang Luong, and Quoc Le.
\newblock Unsupervised data augmentation for consistency training.
\newblock {\em Advances in Neural Information Processing Systems},
  33:6256--6268, 2020.

\bibitem{ye2019unsupervised}
Mang Ye, Xu Zhang, Pong~C Yuen, and Shih-Fu Chang.
\newblock Unsupervised embedding learning via invariant and spreading instance
  feature.
\newblock In {\em Proceedings of the IEEE/CVF Conference on Computer Vision and
  Pattern Recognition}, pages 6210--6219, 2019.

\bibitem{yu2022adversarial}
Qiying Yu, Jieming Lou, Xianyuan Zhan, Qizhang Li, Wangmeng Zuo, Yang Liu, and
  Jingjing Liu.
\newblock Adversarial contrastive learning via asymmetric infonce.
\newblock In {\em European Conference on Computer Vision}, pages 53--69.
  Springer, 2022.

\bibitem{zhai2019adversarially}
Runtian Zhai, Tianle Cai, Di He, Chen Dan, Kun He, John Hopcroft, and Liwei
  Wang.
\newblock Adversarially robust generalization just requires more unlabeled
  data.
\newblock {\em arXiv preprint arXiv:1906.00555}, 2019.

\bibitem{zhang2019theoretically}
Hongyang Zhang, Yaodong Yu, Jiantao Jiao, Eric Xing, Laurent El~Ghaoui, and
  Michael Jordan.
\newblock Theoretically principled trade-off between robustness and accuracy.
\newblock In {\em International conference on machine learning}, pages
  7472--7482. PMLR, 2019.

\bibitem{zhang2019making}
Richard Zhang.
\newblock Making convolutional networks shift-invariant again.
\newblock In {\em International conference on machine learning}, pages
  7324--7334. PMLR, 2019.

\end{thebibliography}
}

\end{document}